\documentclass{bmvc2k}

\title{Scene Parameter Saliency via Differentiable Light Transport}

\addauthor{Linas Beresna}{linas_beresna@sfu.ca}{1}
\addauthor{Eugene Fiume}{eugene_fiume@sfu.ca}{1}

\addinstitution{
 Simon Fraser University\\
 Burnaby, Canada
}

\runninghead{Linas Beresna and Eugene Fiume}{Scene Parameter Saliency}

\def\eg{\emph{e.g}\bmvaOneDot}

\def\etal{\emph{et al}\bmvaOneDot}

\usepackage{amsmath}
\usepackage{bm}
\usepackage{amssymb}
\usepackage{graphicx}
\usepackage{wrapfig}
\usepackage{caption}

\begin{document}

\maketitle

\begin{abstract}
Gradient-based saliency methods reveal which input features most influence a neural network's output, and are a standard tool for model interpretability. We observe that differentiable renderers, which are conventionally used for parameter optimisation, produce an analogous form of saliency: given any scalar metric evaluated on a rendered image, a single reverse-mode differentiation pass yields per-parameter gradients that identify which scene elements most influence the metric. We call these gradient fields metric saliency maps. Unlike neural saliency, which propagates attribution through learned weights, metric saliency propagates through the image formation process itself, including multi-bounce light transport, capturing parameter dependencies that are semi-opaque to manual inspection. We compute metric saliency maps for qualitatively different objectives: psychovisual glare indices, mean scene luminance, and neural perceptual scores. The saliency rankings differ substantially across metrics for the same scene, with parameters that dominate one objective being negligible for another. The saliency map is specific to the metric, not an intrinsic property of the scene. Our results suggest that differentiable renderers produce derivative images that are as informative for scene understanding as the primal images they were designed to generate.
\end{abstract}

\section{Introduction}
\label{sec:intro}
 
A central concern of computer vision is understanding how images relate to the scenes that produce them. Classical physics-based vision tackled this relationship directly: shape from shading~\cite{Horn1989Shading}, intrinsic image decomposition~\cite{Barrow1978Recovering}, and photometric stereo~\cite{Woodham1980Photometric} all reason about the physical process of image formation to recover scene properties from pixel observations. These methods succeed because the mapping from scene to image, governed by geometry, reflectance, and illumination, is structured and (in principle) invertible.
 
Differentiable rendering~\cite{Li2018Differentiable, Mitsuba3, Laine2020Modular} has revived this physics-based perspective with modern computational tools. By implementing the rendering pipeline as an automatically differentiable program, one obtains gradients of any image-derived quantity with respect to scene parameters: geometry, materials, lighting, and camera. These gradients have been used extensively for inverse rendering~\cite{Li2018Differentiable, NimierDavidVicini2019Mitsuba2}, pose estimation~\cite{Tremblay2023Diff}, shape reconstruction~\cite{Liu2019Soft, Chen2019Learning}, and scene generation~\cite{Yuan2024Diffcsg}. In all of these applications, the gradient serves a single purpose: it is fed into an optimiser that iteratively updates scene parameters to minimise a loss function.
 
We observe that these gradients carry information that is valuable \emph{independently of any optimisation}. Given a scalar metric $M$ evaluated on a rendered image and a differentiable renderer $\mathcal{R}$ parameterised by scene parameters $\boldsymbol{\theta}$, the gradient $\partial M(\mathcal{R}(\boldsymbol{\theta})) / \partial \boldsymbol{\theta}$ is a vector that assigns to every scene parameter a score reflecting how much it influences the metric. This is a saliency map over the parameter space of the scene.
 
The analogy to neural network interpretability is direct. Gradient-based saliency methods such as Grad-CAM~\cite{Selvaraju2017Grad} and vanilla gradient attribution~\cite{Simonyan2013Deep} reveal which input features most influence a network's prediction by inspecting $\partial y / \partial \mathbf{x}$. Our metric saliency maps do the same, but the differentiable function is a physically-based renderer rather than a learned model, and the ``input features'' are scene parameters rather than pixel values. Where Grad-CAM asks \emph{which pixels matter for this classification?}, we ask \emph{which scene parameters matter for this metric?}

\begin{figure}[t]
  \centering
  \small
  \includegraphics[width=\linewidth, trim={0 0 0 0}, clip]{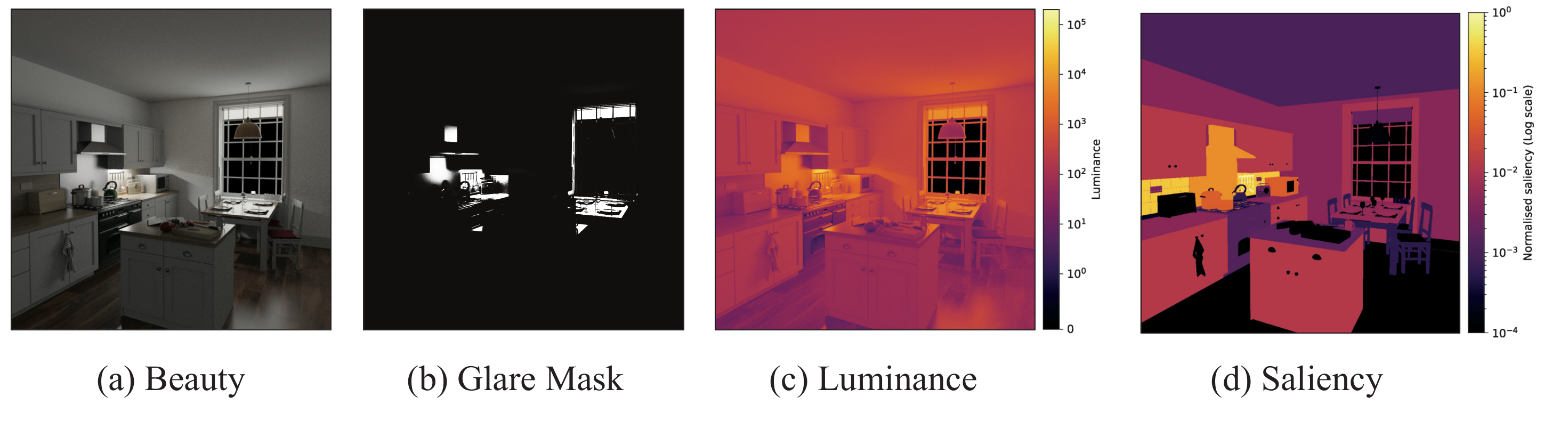}
  \vspace{0em}
  \caption{\textbf{Expanding the diagnostic toolkit for architectural lighting.} While conventional analysis provides (b) glare source identification and (c) luminance maps to evaluate discomfort glare, it leaves the designer to guess the root physical causes. By differentiating the psychovisual unified glare rating through the light transport simulation, we obtain (d) a metric saliency map. This highlights the specific scene materials and geometry driving the metric, explicitly guiding design decisions.}
  \label{fig:teaser}
  \vspace{-3mm}
\end{figure}
 
The distinction between neural and physical saliency is more than cosmetic. Neural saliency propagates through learned weights whose structure is opaque by nature. Metric saliency propagates through the image formation process itself, including multi-bounce light transport, where the dependencies have a physical interpretation. A wall surface three bounces removed from a light source may, through interreflection, have a larger effect on a glare metric than the luminaire itself. The gradient encodes this dependency exactly, in a single reverse-mode differentiation pass, without requiring the repeated evaluations of parametric sweeps~\cite{Sobol2001Global} or finite-difference probing that characterise traditional sensitivity analysis in architectural and lighting design~\cite{Heschong2002Daylighting, Reinhart2019Daylight}.
 
We demonstrate metric saliency maps across qualitatively different objectives, including mean scene luminance, neural perceptual similarity scores, and psychovisual glare ratings. Figure \ref{fig:teaser} illustrates this approach for architectural lighting by explicitly mapping the root physical causes of discomfort glare. The saliency rankings differ substantially across metrics for the same scene. Parameters that dominate one objective are negligible for another, confirming that the saliency map is specific to the metric rather than being an intrinsic property of the scene. This metric specificity is the practical value of the tool because it tells a designer not just \emph{what to change}, but \emph{what matters for this particular criterion}.
 
Our contributions are:
\begin{enumerate}
    \item We identify metric saliency maps, the per-parameter gradients of a scalar objective through a differentiable renderer, as an interpretability tool analogous to gradient-based attribution in neural networks, and distinct from the optimisation role these gradients conventionally serve.
    \item We demonstrate that a single reverse-mode differentiation pass through physically-based rendering exposes non-obvious parameter sensitivities across diverse metrics, including cases where the most influential element is spatially and optically distant from the measured effect.
    \item We show that saliency rankings are metric specific: changing the objective function over the same scene produces qualitatively different maps, providing diagnostic information that neither the primal image nor any single metric value can convey alone.
\end{enumerate}
 
\section{Related Work}
\label{sec:related}
 
Our work sits at the intersection of differentiable rendering, gradient-based interpretability, and physics-based scene analysis. We survey each area and identify the gap that metric saliency maps address.
 
\paragraph{Differentiable rendering.}
The ability to compute image gradients with respect to scene parameters has been developed across rendering paradigms. For rasterisation, Neural Mesh Renderer~\cite{Kato2018Neural}, SoftRas~\cite{Liu2019Soft}, DIB-R~\cite{Chen2019Learning}, and nvdiffrast~\cite{Laine2020Modular} introduce differentiable approximations to the discrete visibility function. For Monte Carlo path tracing, Li~\etal~\cite{Li2018Differentiable} proposed edge sampling for boundary gradients, Loubet~\etal~\cite{Loubet2019Reparameterizing} introduced reparameterisation-based estimators, and Nimier-David~\etal~\cite{NimierDavid2020Radiative} developed radiative backpropagation within Mitsuba~2. Vicini~\etal~\cite{NimierDavidVicini2019Mitsuba2} and the Path Replay Backpropagation (PRB) method~\cite{Vicini2021PathReplayBackprop} further reduced the memory and variance costs of differentiating through global illumination. Mitsuba~3~\cite{Mitsuba3} consolidates many of these techniques into a single framework. Neural scene representations, including NeRF~\cite{mildenhall2021nerf} and 3D Gaussian Splatting~\cite{kerbl20233d}, are also inherently differentiable and have been used for view synthesis and reconstruction~\cite{barron2021mip, gao2024relightable}. DiffCSG~\cite{Yuan2024Diffcsg} extends differentiability to constructive solid geometry and notably includes per-parameter gradient visualisations as part of its pipeline. Across all of these systems, gradients are treated as an intermediate quantity consumed by an optimiser. We propose to treat them as a final output consumed by a human.
 
\paragraph{Gradient-based saliency and attribution.}
In the neural network interpretability literature, gradient-based methods are a primary tool for explaining model decisions. Simonyan~\etal~\cite{Simonyan2013Deep} introduced vanilla gradient saliency, computing $\partial y / \partial \mathbf{x}$ to highlight input pixels that most affect a classifier's output. Grad-CAM~\cite{Selvaraju2017Grad} and Grad-CAM++~\cite{chattopadhay2018grad} refine this by weighting feature map activations with gradients, producing class-discriminative heatmaps. These methods have been extended to object detection, segmentation, and visual question answering~\cite{boggust2023saliency}. Integrated Gradients~\cite{sundararajan2017axiomatic} and SmoothGrad~\cite{smilkov2017smoothgrad} address gradient noise and saturation. The general principle, that the derivative of a differentiable function with respect to its inputs constitutes an attribution over those inputs, is well established for neural networks. We apply this same principle to a different differentiable function: a physically-based renderer. The ``inputs'' become scene parameters (materials, geometry, lighting) rather than pixel values, and the ``output'' becomes an arbitrary scalar metric rather than a class logit.
 
\paragraph{Physics-based vision and scene understanding.}
Our work connects to a tradition in computer vision that reasons about image formation through physical models. Shape from shading~\cite{Horn1989Shading}, intrinsic image decomposition~\cite{Barrow1978Recovering, barron2014shape}, and photometric stereo~\cite{Woodham1980Photometric} all invert the rendering process to recover scene properties. More recently, analysis-by-synthesis approaches use differentiable rendering to optimise scene explanations that match observed images~\cite{yuille2006vision}. MRD~\cite{beilharz2025mrd} uses physically-based differentiable rendering to probe vision models by finding scene-parameter metamers, physically different scenes that produce identical model activations, offering an interpretability tool grounded in 3D scene descriptions rather than pixel space. Our approach shares MRD's motivation of using differentiable rendering for understanding rather than reconstruction, but operates through gradient inspection rather than optimisation, and targets scalar scene metrics rather than neural network representations.
 
\paragraph{Sensitivity analysis in design.}
In architectural lighting and building performance evaluation, understanding how design parameters affect metrics such as Unified Glare Rating (UGR)~\cite{cie117_1995}, Daylight Glare Probability (DGP)~\cite{wienold2006evaluation}, spatial daylight autonomy~\cite{lm2013approved}, and useful daylight illuminance~\cite{nabil2006useful} is essential for informed design. Traditional approaches rely on parametric sweeps~\cite{Heschong2002Daylighting}, variance-based sensitivity analysis (Sobol indices)~\cite{Sobol2001Global}, or Morris screening~\cite{morris1991factorial}, all of which require many independent simulation evaluations and scale poorly with the number of parameters. Gradient-based lighting optimisation has been explored using differentiable rendering for luminaire placement~\cite{lipp2024View, savage2026differentiable}, 
but the gradients serve an optimiser rather than a human. Our metric saliency maps provide a local sensitivity characterisation at the cost of a single backward pass.

\section{Method}
\label{sec:method}
 
The method has three stages: render an image through a differentiable pipeline, evaluate a scalar metric on that image, and backpropagate through both the metric and the renderer to obtain per-parameter gradients. These gradients are then mapped back to image space for visualisation. We describe each stage and the metrics we evaluate.
 
\subsection{Differentiable Image Formation}
\label{sec:method:pipeline}
 
Let $\mathcal{R}$ be a renderer that computes an image $I = \mathcal{R}(\boldsymbol{\theta})$ from scene parameters $\boldsymbol{\theta}$. The parameter vector $\boldsymbol{\theta}$ may include surface reflectances, roughness values, indices of refraction, emitter intensities, and geometric quantities. We use a physically-based Monte Carlo path tracer with multi-bounce global illumination, so that each pixel is a function of parameters potentially distributed across the entire scene through interreflection. Given a scalar metric $M$ evaluated on the rendered image, the key requirement is that the full composition $M(\mathcal{R}(\boldsymbol{\theta}))$ is differentiable: both the renderer $\mathcal{R}$ and the metric $M$ must admit gradient computation, so that we can obtain $\partial M / \partial \boldsymbol{\theta}$ via reverse-mode automatic differentiation through the metric and the light transport simulation jointly. We use Mitsuba~3~\cite{Mitsuba3} with PRB~\cite{Vicini2021PathReplayBackprop}, though any differentiable renderer would serve the same role.
 
\subsection{Metric Saliency Computation}
\label{sec:method:saliency}
 
Given a scalar measurement operator $M : \mathbb{R}^D \rightarrow \mathbb{R}$ that evaluates a property of the light transport simulation, let $I = \mathcal{R}(\boldsymbol{\theta}) \in \mathbb{R}^D$ represent a generalised sensor response. By expanding the traditional image formation model, $I$ can encode standard visible images (where $D = H \times W \times C$) as well as non-visible or abstract radiometric properties. The composition $M(\mathcal{R}(\boldsymbol{\theta}))$ thus defines a scalar function of the scene parameters. Applying the chain rule yields the parameter-space gradient:
\begin{equation}
    \mathbf{g} = \frac{\partial M(\mathcal{R}(\boldsymbol{\theta}))}{\partial \boldsymbol{\theta}} = \frac{\partial M}{\partial I} \frac{\partial I}{\partial \boldsymbol{\theta}}.
    \label{eq:gradient}
\end{equation}
The chain rule separates this into two terms: the sensitivity of the metric to the image, which depends on how $M$ is defined, and the sensitivity of the image to scene parameters, which depends on the light transport. Neither term needs to be computed or inspected independently; the product is obtained automatically by backpropagation.
 
Each component $g_i = \partial M / \partial \theta_i$ has a direct interpretation: it measures how much $M$ would change per unit perturbation of $\theta_i$, accounting for all direct and indirect light paths. The magnitude $|g_i|$ indicates importance, the sign indicates direction. We define the \emph{saliency} of parameter $\theta_i$ as $|g_i|$, and for vector-valued parameters (\eg, an RGB reflectance) we take the $\ell_2$ norm of the gradient components.
 
\subsection{Image-Space Visualisation}
\label{sec:method:visualisation}

\begin{wrapfigure}{r}{0.45\textwidth}
  \vspace{-5em}
  \centering
  \includegraphics[width=\linewidth, trim={0 0 0 0}, clip]{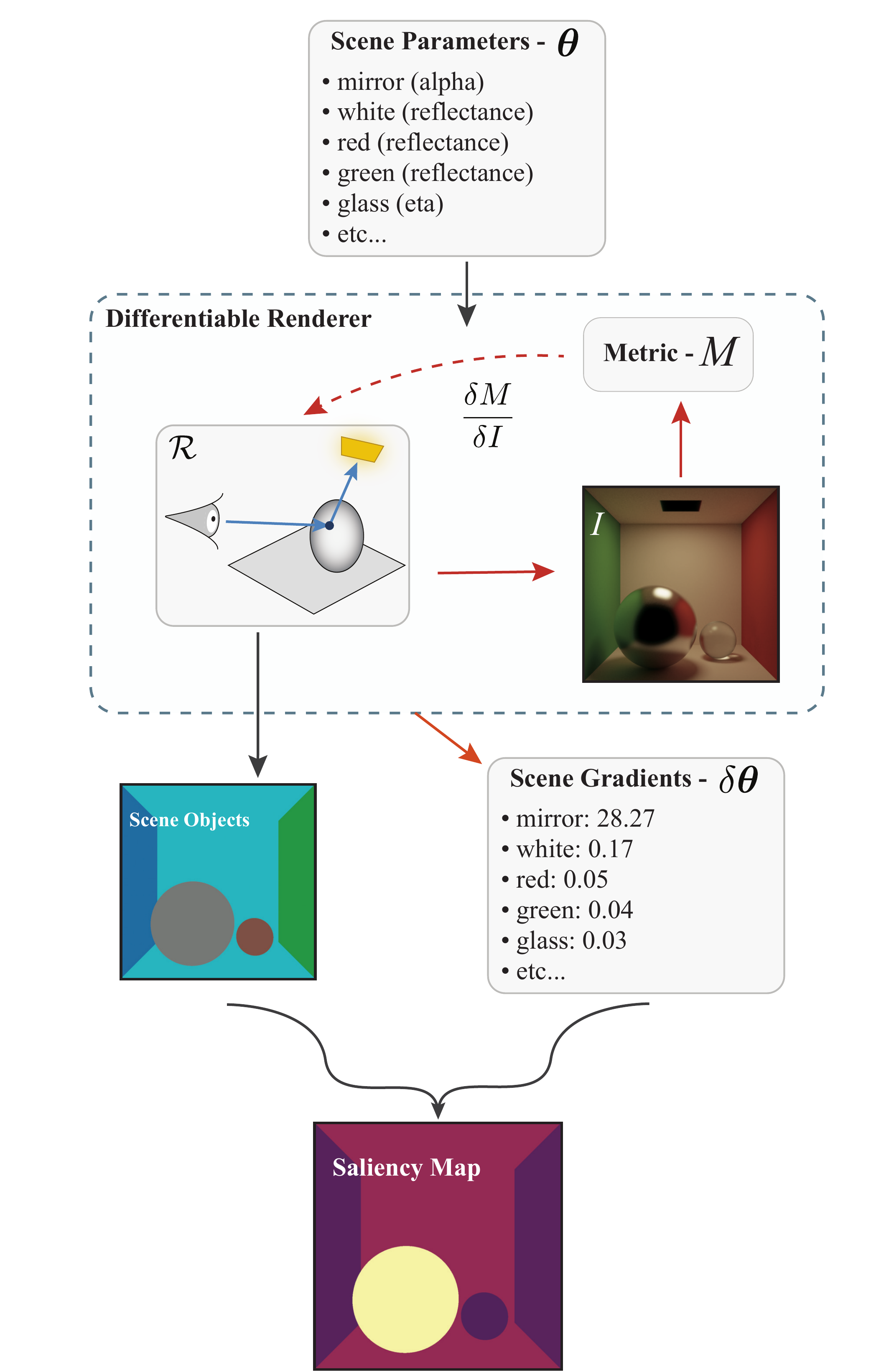}
  \vspace{0.5em}
  \caption{\textbf{Image-Space Saliency Mapping.} Per-pixel material identities (Scene Objects) are coloured by their corresponding parameter sensitivities (Scene Gradients) to produce the final metric-specific heatmap.}
  \label{fig:pipeline}
  \vspace{-1.5em}
\end{wrapfigure}

As illustrated in Figure~\ref{fig:pipeline}, the gradient vector $\mathbf{g}$ lives in parameter space: each component corresponds to a named scene parameter, not a pixel. To produce an image-space visualisation, we need to associate each pixel with the scene parameter(s) visible at that pixel. We achieve this by tracing primary rays from the camera and, at each intersection, identifying which scene object (and therefore which material parameters) the ray hit. This produces a per-pixel material identity map, analogous to an instance segmentation mask in computer vision or a cryptomatte pass in visual effects compositing~\cite{friedman2015fully}. Each pixel is then coloured by the saliency of the material visible at that location, producing a heatmap that can be compared side-by-side with the beauty render.

For scalar parameters that apply uniformly to an entire object (\eg, a single roughness value for a surface), every pixel showing that object receives the same saliency value. For spatially-varying parameters such as texture maps, the gradient is itself spatially varying and can be mapped directly onto the surface. In this work we focus on uniform material parameters; spatially-varying saliency fields are left to future work.

We normalise saliency values across all parameters to $[0, 1]$ for visualisation, so that the map highlights \emph{relative} importance within a scene configuration. The resulting image is metric specific: the same scene rendered under the same conditions will produce different saliency maps for different metrics. 
 
\subsection{Choice of Metrics}
\label{sec:method:metrics}
 
To demonstrate generality, we evaluate metrics from three qualitatively different families.
 \vspace{-1em}
\paragraph{Mean luminance.} This is the simplest metric: the average luminance of the rendered image, computed as a weighted sum of RGB channels using standard photometric coefficients. This serves as a baseline. The saliency map should highlight surfaces with large visible area and high reflectance, which are the parameters that most affect overall image brightness.
 \vspace{-1em}
\paragraph{Unified Glare Rating (UGR).} A psychovisual metric standardised by the CIE~\cite{cie117_1995} for evaluating discomfort glare in architectural lighting. UGR aggregates the luminance, solid angle, and positional weight of glare sources against a background luminance. It is non-linear, involves a logarithm and a ratio, and its dependence on scene parameters is mediated by the glare/background decomposition and the Guth position index~\cite{guth1966computing}.
\vspace{-1em}

 
\paragraph{Neural perceptual score.} To demonstrate that the approach extends beyond hand-crafted metrics, we evaluate a metric defined by a neural network. Specifically, we backpropagate from the pre-softmax class logit of a pre-trained ResNet-50 image classifier \cite{he2016deep}. This case is notable because the gradient must flow through both the neural network latent space and the physical light transport simulation, and the resulting saliency map reflects the joint sensitivity of both systems.

\section{Results}
\label{sec:results}

\paragraph{Experimental Setup.} We render standard interior scenes including a dining room, kitchen, and living room using PRB~\cite{Vicini2021PathReplayBackprop} with Mitsuba 3~\cite{Mitsuba3} and DrJit~\cite{Jakob2020DrJit}. We trace paths up to a maximum depth of six bounces with 1024 samples per pixel. For each scene, we track all differentiable material parameters and evaluate metric saliency for mean luminance, discomfort glare, and neural classification logits.

\subsection{Metric Specificity}
\label{sec:results:specificity}

\begin{figure}[t]
  \centering
  \small
  \vspace{1em}
  \includegraphics[width=\linewidth, trim={0 0 0 0}, clip]{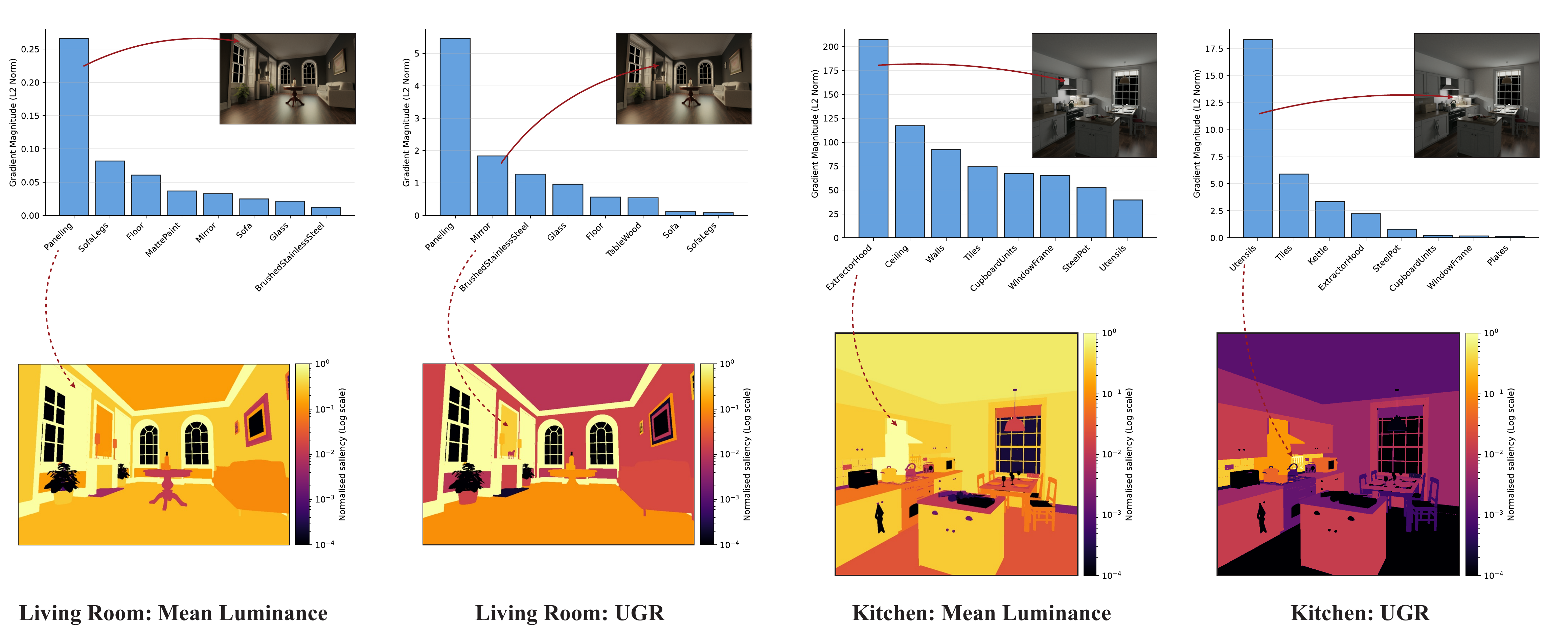}
  \vspace{1em}
  \caption{\textbf{Visualizing Metric Specificity.} While extracting gradients provides precise ranking data for scene parameters (top row), mapping this raw data back into the 3D scene (bottom row) allows for intuitive spatial visualisation. Across multiple scenes, the UGR evaluation isolates highly localised reflective surfaces driving discomfort glare (e.g., the mirror or metallic utensils). In contrast, the mean luminance evaluation shifts focus entirely to large surfaces driving global illumination.}
  \label{fig:comparison}
  \vspace{-2em}
\end{figure}

Psychovisual metrics like UGR depend on complex spatial relationships between the observer and the environment, whereas simple metrics like mean luminance measure global scene energy. Figure \ref{fig:comparison} demonstrates how our pipeline adapts to these fundamentally different objectives across both a living room and a kitchen environment.

By default, the differentiable engine produces a raw mathematical list of parameter sensitivities. While plotting these gradients as bar charts (top row) yields precise statistical rankings, isolating this data from its spatial context makes it difficult for a human designer to parse. By aggregating these parameter gradients and mapping them directly back onto the 3D scene geometry, we produce an intuitive visual heatmap of metric sensitivity (bottom row).

The UGR saliency maps explicitly isolate the specific reflective surfaces and light sources that maximise discomfort glare for the given camera pose, such as the mirror in the living room or the metallic utensils in the kitchen. This provides highly localised, actionable targets for a designer to mitigate visual discomfort. To verify that this is not simply a generic scene feature map, we compare it against the mean luminance objective. As shown in Figure \ref{fig:comparison}, the parameters dictating overall scene brightness differ from those causing localised glare. The mean luminance maps correctly shift emphasis to large geometric areas, such as the diffuse walls, ceilings, and floors, confirming that our spatial projection accurately visualises the underlying physical drivers of the chosen metric regardless of the scene's architecture.

\begin{wrapfigure}{r}{0.45\textwidth}
  \vspace{-9em}
  \centering
  \includegraphics[width=\linewidth, trim={0 0 2.0cm 0}, clip]{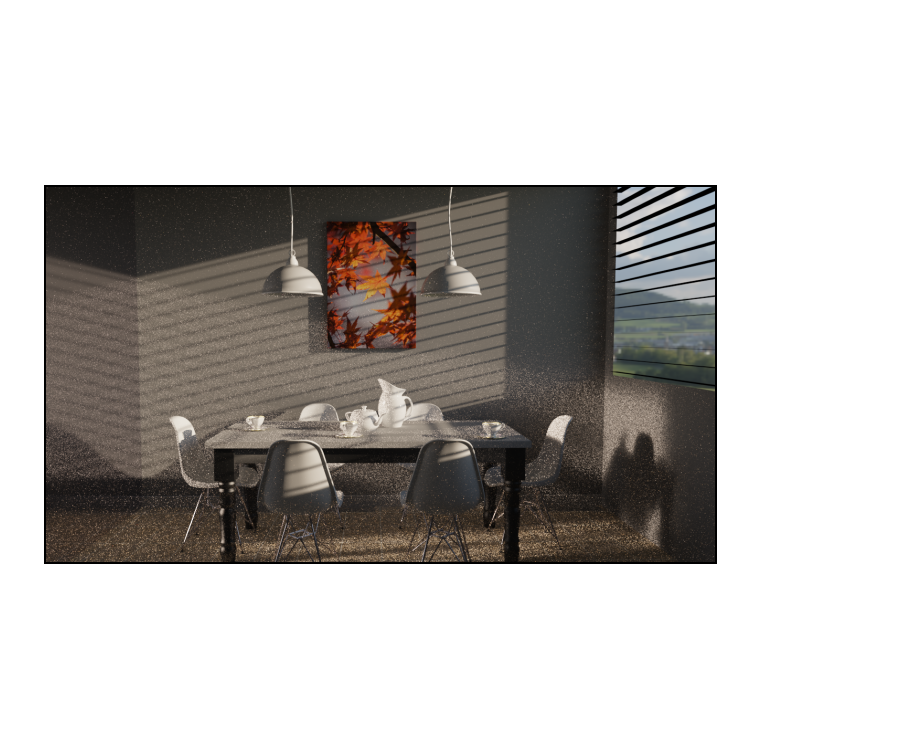}\vspace{-3.9em}
  {\small (a) Beauty Render}\vspace{-3em}
  \includegraphics[width=\linewidth, trim={0 0 2.0cm 0}, clip]{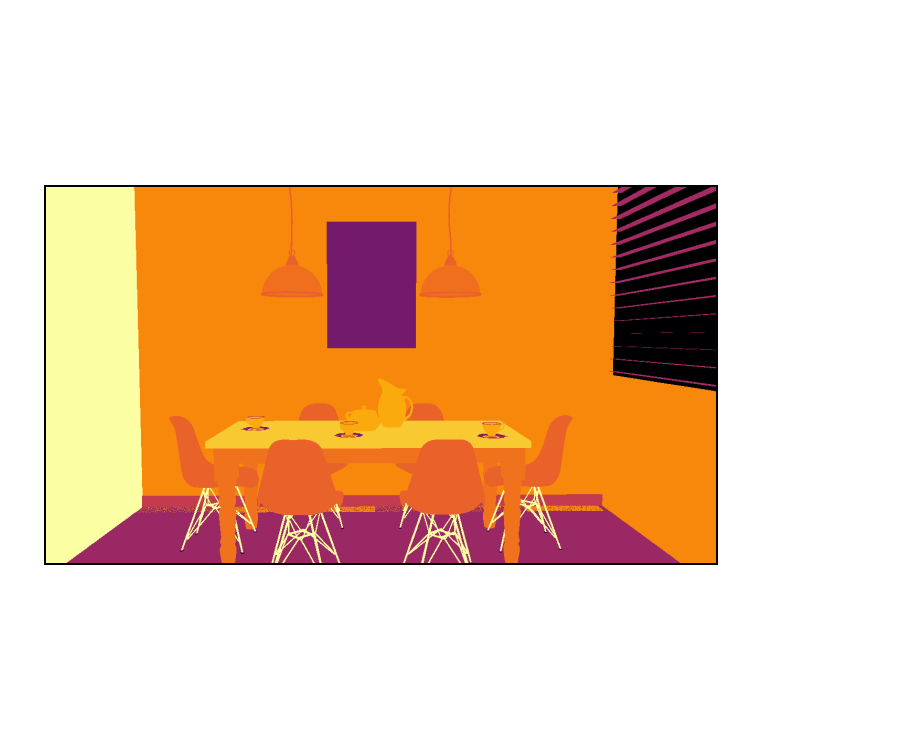}\vspace{-3.9em}
  {\small (b) Saliency Map}
  \vspace{1em}
  \caption{\textbf{Semantic attribution mapped into 3D parameter space.} The saliency map shows which materials act as the cues for the classification logit.}
  \label{fig:resnet_results}
  \vspace{-2em}
\end{wrapfigure}

\subsection{Neural and Perceptual Scores}
\label{sec:results:neural}

We demonstrate that this pipeline handles high dimensional learned spaces by evaluating a semantic classifier. Figure \ref{fig:resnet_results} illustrates the metric saliency for the dining room scene when backpropagating from the ResNet-50 logit for the folding chair class (class 559).

The gradient flows from the neural network latent space directly into the physical light transport simulation, capturing complex optical dependencies. Notably, the left wall of this scene consists of a highly reflective chrome material. The resulting saliency map explicitly visualises how this specular surface influences the classification score. It highlights that the neural network relies not only on direct geometric edges, but also on the reflected features and specific material roughness values to conceptually recognise the objects in the scene.

\subsection{Validation via Independent Parameter Ablation}
\label{sec:results:validation}

\begin{figure}[h]
  \centering
  \small
  \begin{tabular}{@{}c@{\hspace{4pt}}c@{}}
    \includegraphics[width=0.48\textwidth]{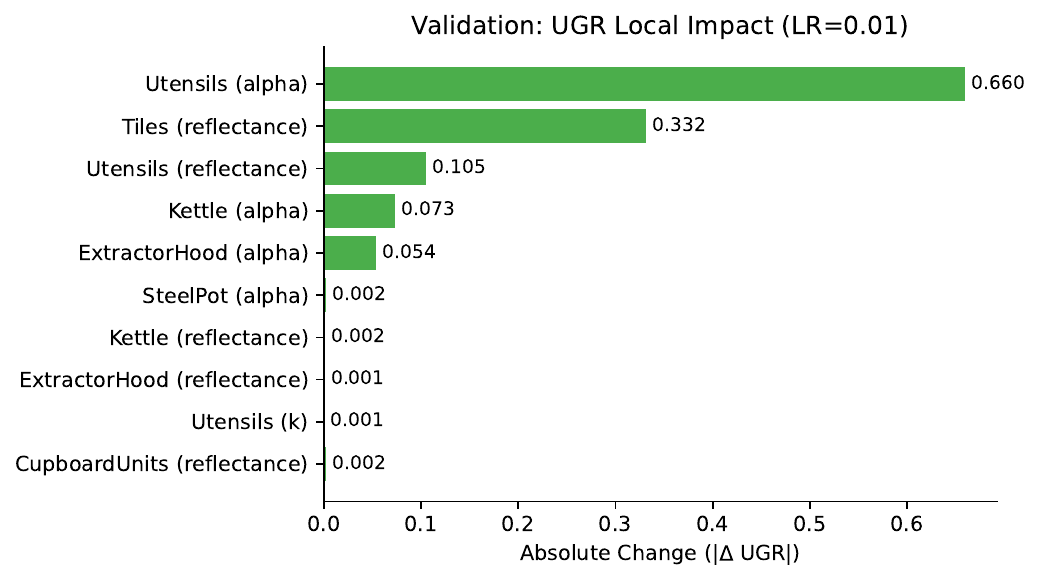} &
    \includegraphics[width=0.48\textwidth]{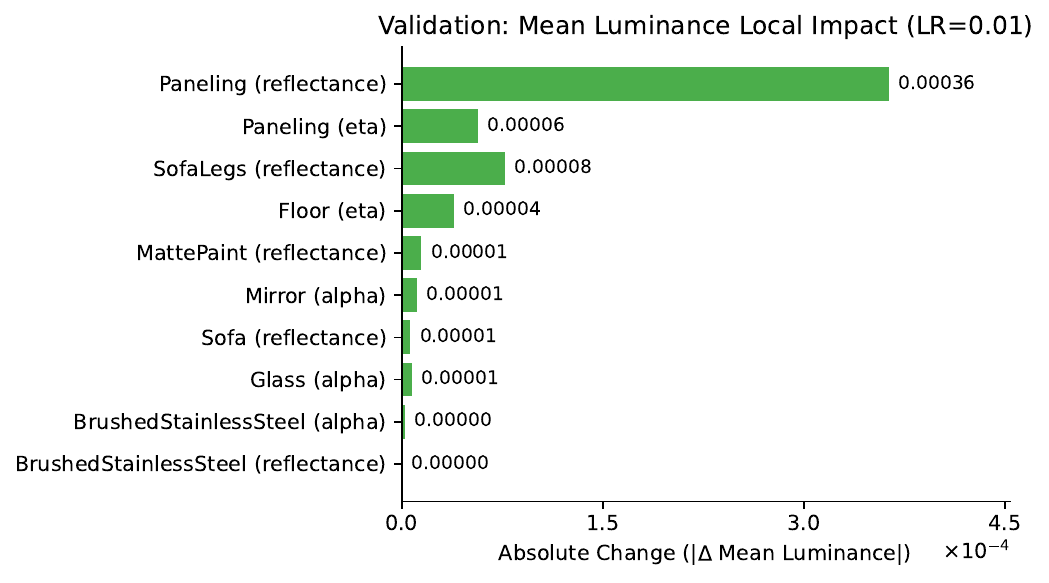} \\
    (a) Kitchen UGR Validation & (b) Living Room Mean Luminance Validation
  \end{tabular}
  \vspace{3mm}
  \caption{\textbf{Independent Ablation Validation.} Parameters are ranked strictly by computed gradient magnitude (y-axis). Taking a small gradient descent step yields an absolute change in the objective metric (x-axis). The UGR evaluation (a) perfectly mirrors the computed saliency rank, whilst the mean luminance evaluation (b) highlights the noise floor challenges of global stochastic metrics.}
  \label{fig:validation_ablation}
\end{figure}

To verify that our computed gradients accurately predict the physical sensitivities of the objective functions, we perform an independent ablation test on both the non-linear UGR metric and the global mean luminance metric (Figure \ref{fig:validation_ablation}). Because these metrics represent complex optical dependencies, a computed gradient provides a strictly local first-order approximation. To rigorously validate this without colliding with physical material bounds (e.g., surface roughness clamping at $1.0$), we evaluate the metric's actual response to a small, independent gradient descent step (learning rate $ = 0.01$) for each of the top-ranked parameters.

As shown in Figure \ref{fig:validation_ablation}a, the absolute measured impact on the localised UGR metric perfectly follows our computed saliency ranking in a monotonically decreasing curve. This confirms that a single reverse-mode evaluation in our differentiable framework successfully isolates and ranks the true physical drivers of discomfort glare, effectively replacing the need for dozens of expensive finite-difference simulations.

When evaluating global metrics like mean luminance (Figure \ref{fig:validation_ablation}b), the highest-ranked parameter correctly dominates the physical impact. However, at lower ranks, we observe minor deviations from strict monotonicity. This highlights an inherent challenge in differentiable path tracing: as absolute metric changes fall below $10^{-4}$ units, the evaluation plunges into the stochastic noise floor of the renderer. Furthermore, applying a uniform gradient step across fundamentally different physical domains, such as diffuse reflectance versus the index of refraction ($\eta$), yields disparate macroscopic behaviors due to the non-linear nature of Fresnel equations. We leave the exploration of parameter-specific adaptive step scaling and second-order Hessian approximations for future work.
\section{Discussion}
\label{sec:discussion}
 
\paragraph{What the maps reveal.}
The central finding of this work is that metric saliency maps expose parameter-metric relationships that are not directly accessible through inspection of the rendered image or the metric value alone. The primal image shows what the scene looks like; the metric returns a single number summarising some property of that image; but neither reveals \emph{why} the number is what it is, or \emph{what} would change it most effectively. The saliency map fills this gap. The fact that saliency rankings differ substantially across metrics for the same scene confirms that this information is genuinely metric-specific and cannot be recovered from any single rendering or evaluation.
 
\paragraph{Limitations.}
Several limitations should be noted. First, Monte Carlo gradient estimators inherit the variance of the underlying path tracer. At low sample counts, saliency values for individual parameters may be noisy. In our experiments, moderate sample counts (64--256 spp) produced stable rankings, but fine-grained spatially-varying saliency fields would require higher sample budgets or variance reduction techniques. Second, the approach requires the metric $M$ to be differentiable. Metrics with hard thresholds or discrete decisions (\eg, binary glare source detection) must be relaxed into smooth approximations. For instance, evaluating UGR within our framework required us to reformulate standard binary visibility checks into smooth sigmoid relaxations. The choice of relaxation can affect the resulting saliency map, and the relationship between the relaxed and the original metric deserves further investigation. Third, gradient magnitude measures local, first-order sensitivity. It does not capture interactions between parameters (second-order effects), nor does it account for parameter constraints that may prevent a suggested change in practice. A parameter with high saliency may not be actionable if it is physically fixed. Higher-order methods, such as Hessian-based analysis~\cite{wang2025stochastic}, could provide richer sensitivity information at greater computational cost.
 
\paragraph{Scope of the rendering model.}
Our demonstration uses a physically-based Monte Carlo path tracer, but the principle applies to any differentiable image formation process. Rasterisation-based differentiable renderers~\cite{Laine2020Modular, Liu2019Soft}, neural radiance fields~\cite{mildenhall2021nerf}, and Gaussian splatting~\cite{kerbl20233d} all produce parameter gradients that could be interpreted as saliency maps. The multi-bounce setting we emphasise is where the approach is most valuable, because it is precisely where human intuition about parameter influence breaks down, but simpler rendering models would work equally well for simpler scenes.
 
\paragraph{Future work.}
This paper establishes the concept of metric saliency maps and demonstrates their utility. Several directions remain open. The visualisation of gradient information in scene and image space is itself a research question: signed maps, vector-field representations, and interactive exploration interfaces could provide richer and more actionable information than the magnitude heatmaps we present here. Real-time or interactive computation of saliency maps would enable designers to explore parameter sensitivities dynamically as they modify a scene, motivating work on efficient differentiable rendering. Finally, coupling saliency maps with optimisation, using the map to guide which parameters to optimise and which to hold fixed, could combine the interpretability of our approach with the automation of existing inverse rendering pipelines.
 
\section{Conclusion}
\label{sec:conclusion}
 
We have shown that differentiable renderers produce gradient information that is valuable as a direct output, not only as an intermediate quantity for optimisation. By backpropagating a scalar metric through the rendering pipeline, we obtain metric saliency maps that reveal which scene parameters most influence the metric. These maps are inexpensive (a single backward pass), general (applicable to any differentiable metric and renderer), and informative (exposing non-obvious dependencies mediated by multi-bounce light transport). The saliency rankings are metric-specific: the same scene produces qualitatively different maps under different objectives. This metric-specificity is the practical contribution, as it provides diagnostic information that neither the rendered image nor the metric value can convey alone. Rendering has always served a single purpose: making the invisible visible, turning the mathematics of light transport into images that humans can see and reason about. Differentiable rendering computes derivatives of that same process, but to date these derivatives have been consumed exclusively by optimisation algorithms, never treated as a primary interpretability object. We suggest that derivative images deserve the same status as the primal images they accompany: they are structured, interpretable, and made for human eyes.
 
\section{Acknowledgements}
We acknowledge with gratitude the funding provided for this research by Simon
Fraser University as well as a Discovery Grant provided by the Natural Sciences and Engineering Research Council of Canada.

Additionally, we would like to thank Benedikt Bitterli \cite{resources16} for providing the rendering resources (including the Cornell Box,
Grey and White Room, Country Kitchen, and The Breakfast Room models).

\bibliography{references}

\end{document}